\documentclass[conference]{IEEEtran}
\usepackage{cite}
\usepackage{amsmath,amssymb,amsfonts}
\usepackage{algorithmic}
\usepackage{graphicx}
\usepackage{textcomp}
\usepackage{xcolor}
\usepackage{hyperref}
\def\BibTeX{{\rm B\kern-.05em{\sc i\kern-.025em b}\kern-.08em
    T\kern-.1667em\lower.7ex\hbox{E}\kern-.125emX}}
\DeclareMathOperator*{\argmin}{arg\,min}
\begin{document}

\title{Multi-Exit Vision Transformer for Dynamic Inference}

\author{\IEEEauthorblockN{Arian Bakhtiarnia, Qi Zhang and Alexandros Iosifidis}
\IEEEauthorblockA{\textit{DIGIT, Department of Electrical and Computer Engineering, Aarhus University, Denmark}\\
\{arianbakh,qz,ai\}@ece.au.dk}
}

\maketitle

\begin{abstract}

Deep neural networks can be converted to multi-exit architectures by inserting early exit branches after some of their intermediate layers. This allows their inference process to become dynamic, which is useful for time critical IoT applications with stringent latency requirements, but with time-variant communication and computation resources. In particular, in edge computing systems and IoT networks where the exact computation time budget is variable and not known beforehand. Vision Transformer is a recently proposed architecture which has since found many applications across various domains of computer vision. In this work, we propose seven different architectures for early exit branches that can be used for dynamic inference in Vision Transformer backbones. Through extensive experiments involving both classification and regression problems, we show that each one of our proposed architectures could prove useful in the trade-off between accuracy and speed.

\end{abstract}

\section{Introduction}
\label{sec:introduction}

Deep neural networks have achieved immense success in recent years \cite{LeCun2015}, however, they commonly consist of many interconnected layers containing millions of parameters which require high computational resources and cause slow inference speed. Dynamic inference methods \cite{2102.04906} allow deep models to modify their computation graph during inference in order to alleviate this problem. One such method is \textit{early exiting} \cite{Scardapane2020, 2106.11208}, leading to \textit{multi-exit architectures}, where early exit \textit{branches} are inserted after intermediate hidden layers of the \textit{backbone} network and provide early results, albeit with less accuracy compared to the final result of the backbone network.

Early exits are useful in computationally restricted settings such as mobile and edge computing, where early results can be used for ``easy'' inputs to save resources. Additionally, multi-exit architectures can be helpful in \textit{anytime prediction} settings where the inference process may be interrupted at any time and the network is expected to provide a response even if it was interrupted before completion. Examples of anytime prediction settings are distributed environments such as edge computing systems and IoT networks, where the latency depends on the communication channels, which means the exact computation time budget is not known beforehand and varies over time. Here, the latest result provided by a multi-exit architecture can be given as output whenever the network is interrupted.

\textit{Vision Transformer} \cite{dosovitskiy2021an} is a recently proposed architecture for computer vision which has since been applied to various problems, such as image classification, object detection, depth estimation, and many more \cite{2101.01169}. To the best of our knowledge, multi-exit Vision Transformer architectures have not yet been studied in the literature, which limits the application of Vision Transformers in mobile and edge computing. In this work, we propose seven different architectures for early exit branches that can be inserted into Vision Transformer backbones. Through extensive experiments on both image classification and crowd counting, the latter being a regression problem, we show that depending on the particular problem at hand, each of these architectures has the potential to be useful in the trade-off between classification accuracy and inference speed. Our code will be made publicly available at \url{https://gitlab.au.dk/maleci/multiexitvit}.

\section{Related Work}
\label{sec:related_work}

\subsection{Multi-Exit Architectures}

A deep neural network (DNN) can be formulated as a function $f(X) = f_L(f_{L - 1}(...f_1(X))) $ where $ X $ is the input, $ L $ is the number of layers in the DNN and $ f_i $ is the differentiable operator at layer $ i $. The output of layer $ i $ is denoted by $ h_i = f_i(h_{i - 1}) $ and $ \theta_i $ refers to the trainable parameters of $ f_i (\cdot) $. The training process for this DNN can be formulated as shown in Equation \eqref{basic_nn_loss} where $ \theta = \bigcup_{i = 1}^L \theta_i $ is the set of all trainable parameters of the DNN, $ \{X_n, y_n\}_{n = 1}^N $ is the set of training samples and $ l(\cdot) $ is a loss function.

\begin{equation}
f^* = \argmin_{\theta} \sum_{n = 1}^N l(y_n, f(X_n))
\label{basic_nn_loss}
\end{equation}

In order to convert a DNN to a multi-exit architecture, an early exit branch $ c_b(h_b) = y_b $ is placed at every selected branch location $ b \in B \subseteq \{1, .., L\} $, where $ c_b $ is the classifier or regressor producing the early result $ y_b $. The schematic illustration of a multi-exit architecture is shown in Figure \ref{fig:early_exit_and_vit} (a). Since there are multiple outputs in a multi-exit architecture, its training procedure is not as straightforward as Equation \eqref{basic_nn_loss}. Three major strategies for training multi-exit architectures exist in the literature \cite{Scardapane2020}. The \textit{classifier-wise} strategy freezes the backbone, meaning the parameters $ \theta $ will not be modified, and trains the branches separately and independent of each other or the backbone. In the \textit{end-to-end} strategy, the loss function $ l_t = l + \sum_{b \in B} \lambda_b l_b $ combines the losses $ l_b $ of the early exit branches with the backbone's loss and trains the entire multi-exit architecture simultaneously. In this strategy, the contribution of the loss of the branch at location $ b $ is captured by weight score $ \lambda_b $. Finally, the \textit{layer-wise} strategy first trains the layers up to and including the first early exit branch. Subsequently, the previous layers are frozen and the rest of the layers up to and including the second branch are trained, and this operation is repeated until the entire backbone has been trained.

In the end-to-end and layer-wise strategies, the number of branches and their placement create trade-offs between the accuracy of different exits. In addition, with the end-to-end strategy, the weight scores introduce new hyper-parameters. In contrast, no trade-offs or new hyper-parameters need to be considered with the classifier-wise strategy. However, since in this case the parameters of the backbone remain unchanged, fewer parameters are affected during the training of the branches. In this work we investigate all three training strategies.

It is important to note that branches placed later in the networks do not necessarily result in a higher accuracy compared to previous branches. We use the term \textit{impractical} in order to refer to such branches, and the term \textit{practical} for branches with a higher accuracy than all previous branches. The usage of impractical branches would not be sensible since earlier branches with a higher accuracy exist.

\subsection{Vision Transformer}

Vision Transformer (ViT) \cite{dosovitskiy2021an} is an adaptation of the Transformer architecture \cite{NIPS2017_3f5ee243} for computer vision problems. At the core of the Transformer is the \textit{self-attention} layer, which takes a sequence $ X = (x_1, \dots, x_n) \in \mathbb{R}^{n \times d} $ as input and outputs the sequence $ Z = (z_1, \dots, z_n) \in \mathbb{R}^{n \times d_v} $, which can be formulated as Equation \eqref{self_attention}, where  $ Q = XW^Q $, $ K = XW^K $ and $ V = XW^V $ are query, key and value matrices, respectively, in which $ W^Q $, $ W^K $ and $ W^V $ are learnable weight matrices \cite{dosovitskiy2021an}. $ d_k = d_q $ are the size of the vectors in query and key matrices.
\begin{equation}
Z = \mathit{softmax} \left(\frac{QK^T}{\sqrt{d_k}} \right)V
\label{self_attention}
\end{equation}

\begin{figure}
\setlength{\fboxrule}{0pt}
\begin{center}
\begin{tabular}{ c }
\fbox{\includegraphics[height=5cm]{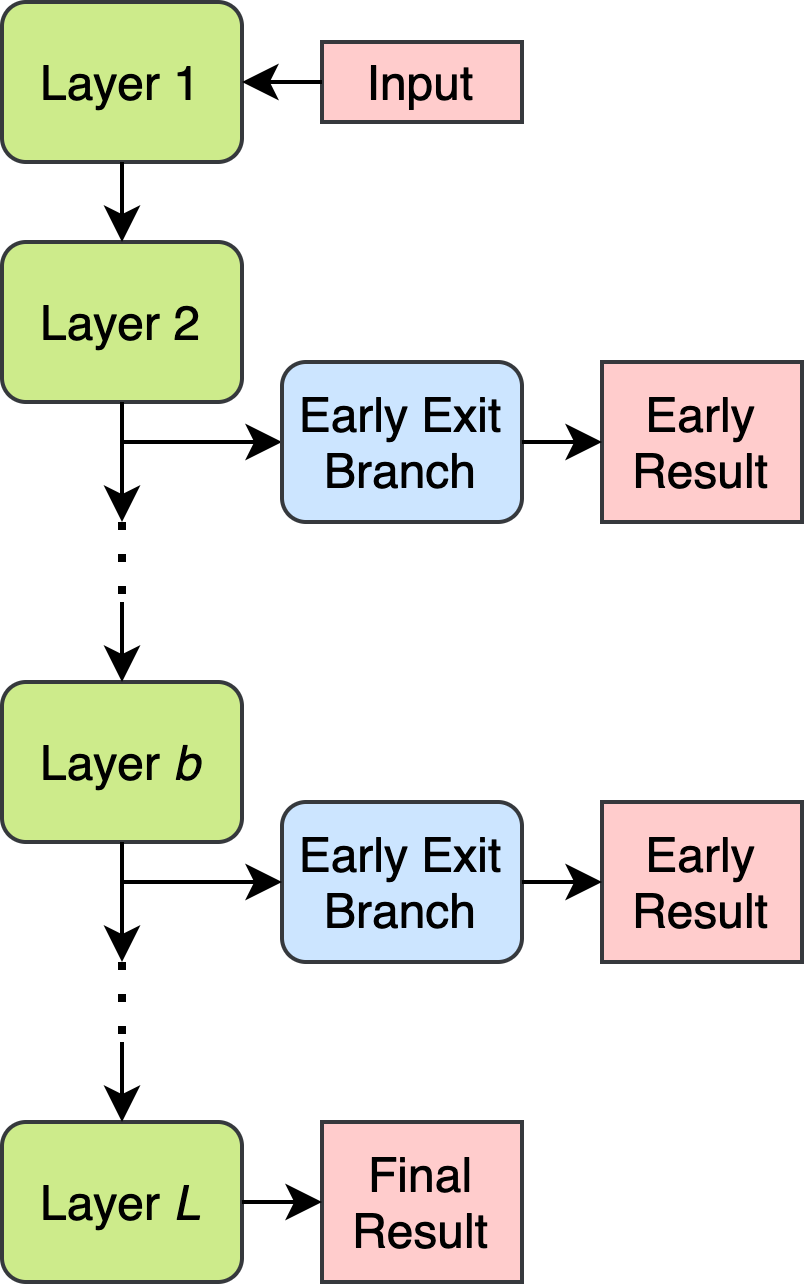}}\\
(a)\\
\fbox{\includegraphics[height=5cm]{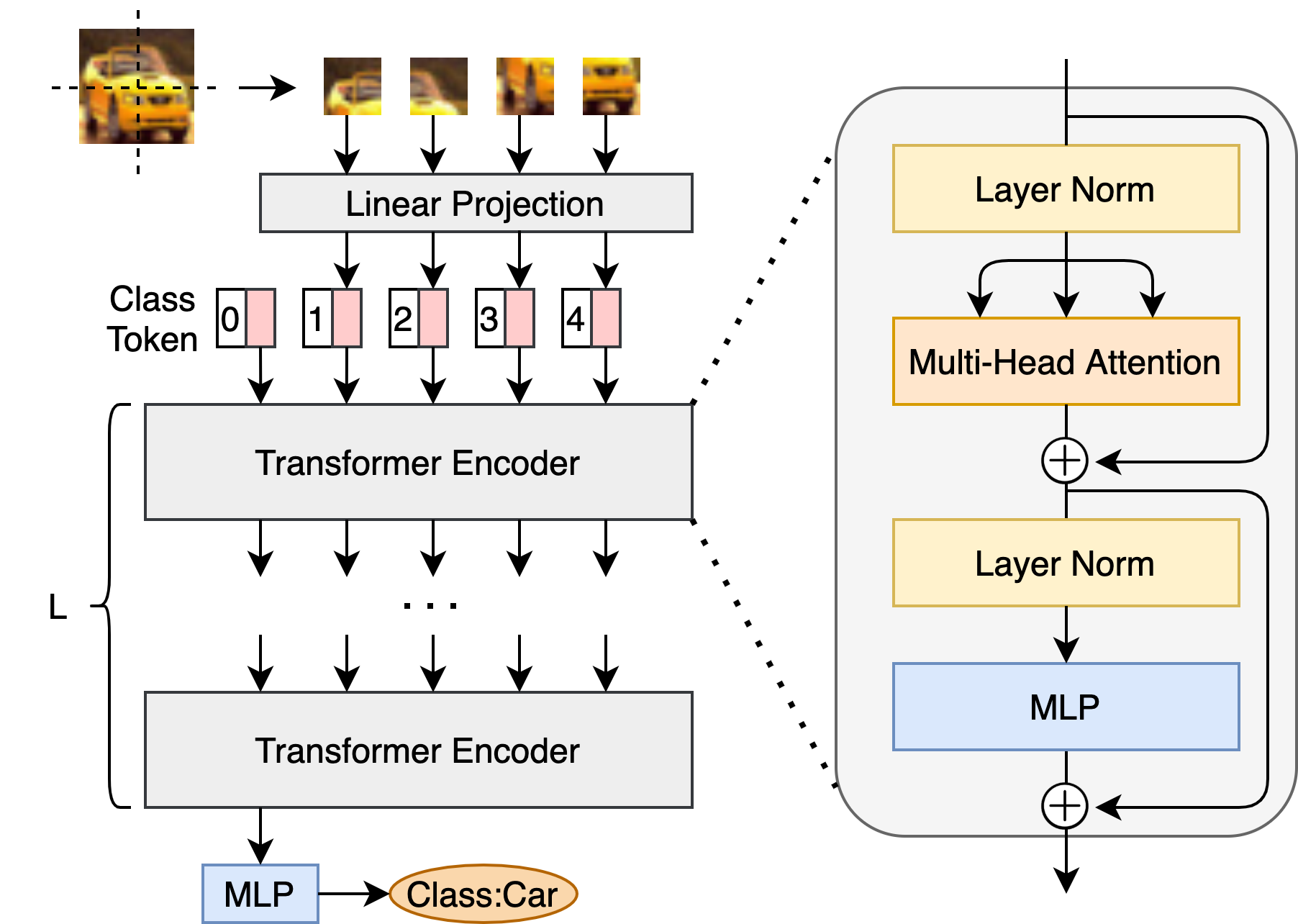}}\\
(b)
\end{tabular}
\end{center}
\caption{(a) Schematic illustration of a multi-exit; and (b) Vision Transformer architecture.}
\label{fig:early_exit_and_vit}
\end{figure}

In order to capture more than one type of relationship between the entities in the sequence, self-attention is extended to \textit{multi-head attention} by concatenating the output of several self-attention blocks, each with its own set of learnable parameters. Figure \ref{fig:early_exit_and_vit} (b) depicts the Vision Transformer architecture, where initially an input image is cut into several image patches. A sequence of patch embeddings is then formed by projecting each patch and concatenating a positional embedding to the resulting vector. An extra learnable \textit{classification token} is also appended to the sequence. The sequence passes through $ L $ Transformer encoder layers, each containing multi-head attention layers among other operations. Finally, the output vector corresponding to the classification token is passed on to an MLP dubbed \textit{classification head} to obtain the final result.

\subsection{Attention-Free, MLP-Based Architectures}

Several MLP-based architectures for computer vision that also operate on sequences of image patches have been recently proposed \cite{2105.15078}. The aim of these architectures is to reduce the computational cost of ViT by removing the attention mechanism, while achieving a comparable performance by preserving a global receptive field similar to that of ViT. Since the intermediate representations in the hidden layers of ViT is in the form of a sequence of patches, it is simple to use the building blocks of these MLP-based architectures as early exit branches placed on ViT backbones. These building blocks create more lightweight branches compared to the Transformer encoders in ViT.

\begin{figure}
\setlength{\fboxrule}{0pt}
\begin{center}
\begin{tabular}{ c }
\fbox{\includegraphics[height=6cm]{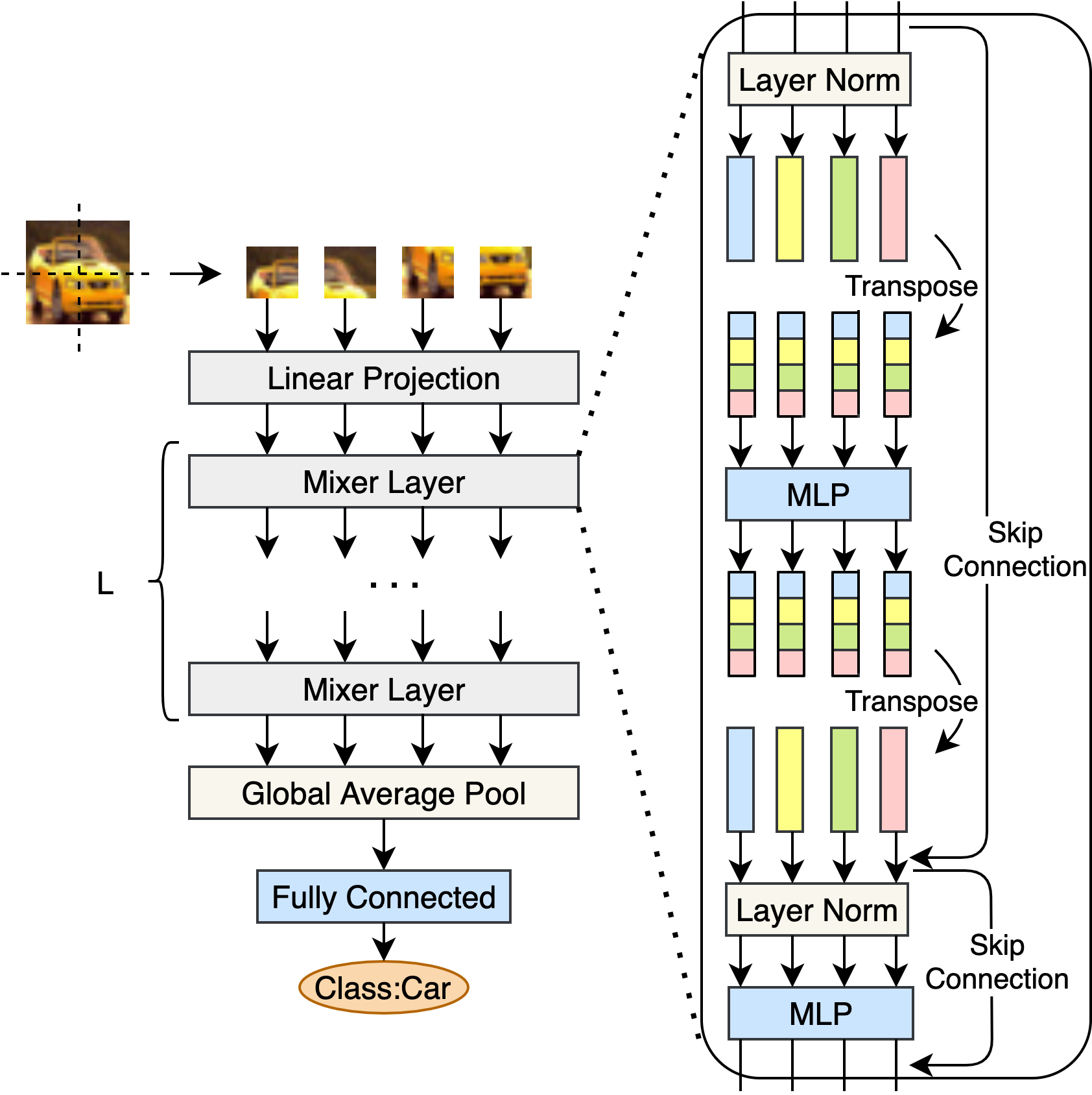}}\\
(a)\\
\fbox{\includegraphics[height=6cm]{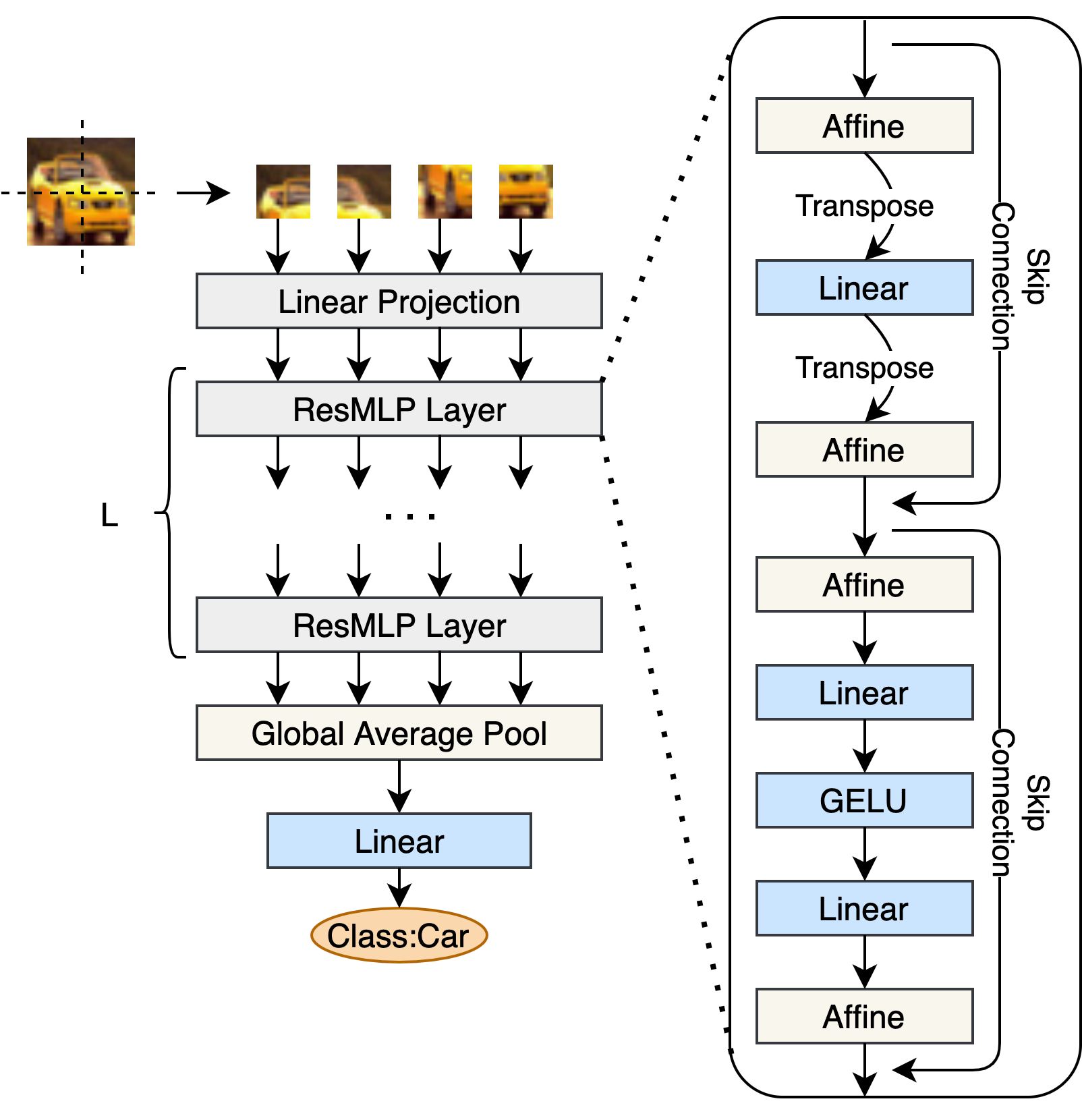}}\\
(b)
\end{tabular}
\end{center}
\caption{(a) MLP-Mixer architecture; and (b) ResMLP architecture.}
\label{fig:mlp_mixer_and_resmlp}
\end{figure}

One such architecture called \textit{MLP-Mixer} \cite{2105.01601} is shown in Figure \ref{fig:mlp_mixer_and_resmlp} (a). Each \textit{mixer layer} in MLP-Mixer consists of \textit{token mixing} and \textit{channel mixing} operations, which are formulated as Equations \eqref{eq:mlp_mixer1} and \eqref{eq:mlp_mixer2}, where $ f_1(\cdot) \dots f_4(\cdot) $ are linear layers and $ \sigma(\cdot) $ is the GELU activation function. The output of the final mixer layer is passed on to a global average pooling layer and then a fully connected layer.
\begin{subequations}
  \begin{align}
    &U = X + f_2(\sigma(f_1(Norm(X)^T)))^T \label{eq:mlp_mixer1} \\
    &Y = U + f_4(\sigma(f_3(Norm(U)))) \label{eq:mlp_mixer2}
  \end{align}
\end{subequations}

A similar architecture called \textit{ResMLP} \cite{2105.03404} is shown in Figure \ref{fig:mlp_mixer_and_resmlp} (b). Each \textit{ResMLP layer} consists of a \textit{cross-patch sublayer} and a \textit{cross-channel sublayer}, which are formulated as Equations \eqref{eq:resmlp1} and \eqref{eq:resmlp2}. In ResMLP, normalization is carried out using an affine transformation instead of layer normalization, as shown in Equation \eqref{eq:resmlp3} where $ \alpha $ and $ \beta $ are learnable vectors that scale and shift the input. Similarly, the output of the final ResMLP layer is passed on to a global average pooling layer and then a fully connected layer.
\begin{subequations}
  \begin{align}
    &U = X + Norm(f_1(Norm(X)^T)^T) \label{eq:resmlp1} \\
    &Y = U + Norm(f_3(\sigma(f_2(Norm(U))))) \label{eq:resmlp2} \\
    &Norm(X) = \mathit{Aff}_{\alpha,\beta}(X) = \mathit{Diag}(\alpha)X + \beta \label{eq:resmlp3}
  \end{align}
\end{subequations}

\section{Multi-Exit Vision Transformer}
\label{sec:multi_exit_vision_transformer}

We assume a high-performing ViT backbone is available for the problem at hand, and the goal is to convert this backbone to a multi-exit architecture in order to allow for dynamic inference. We propose seven different architectures for the early exit branches added after intermediate layers of a ViT backbone. The most intuitive approach, which we call \textit{MLP-EE}, is to add an MLP to the classification token of the intermediate layer, similar to the classification head in the ViT backbone. Even though MLP-EE is very lightweight, it may not contain enough parameters and layers to extract useful features, particularly for exits placed early. Moreover, it does not process tokens other than the classification token.

Another approach is to convert the sequence of token vectors in the intermediate layers of the ViT backbone to a 2D grid and further process them using convolutional filters, leading to 3 different architectures we call \textit{CNN-Ignore-EE}, \textit{CNN-Add-EE} and \textit{CNN-Project-EE}, each handling the classification token in a different way. Note that even though the intermediate layer is in the form of a sequence, each vector in the sequence corresponds to a patch of the input image, therefore putting the vectors back in a 2D grid simulates their original neighborhood which is essential when using convolutional filters that have a local receptive field. The motivation behind this approach is that convolutional filters are the current approach in the literature for early exiting \cite{DBLP:conf/iclr/HuCWW20, Scardapane2020, 7900006} and can act as a baseline for the other proposed architectures. Furthermore, convolutional filters introduce low overhead in terms of parameters and computation. Additionally, a fusion of CNNs that can capture local structure very well but can not handle long range interactions, with ViTs which can process long range interactions, seems natural and may combine the advantages of both \cite{2105.15078}.

On the other hand, the local receptive field of CNN-based early exits may prove to be a drawback. An alternative that can overcome this limitation is using the Transformer encoder layer instead of the convolutional filters, which we call \textit{ViT-EE}. Indeed, it has been shown that Transformer encoder layers can create superior early exits for CNN backbones by introducing a global receptive field \cite{2105.09121}. However, since the layers of ViT backbones already have a global receptive field, it is not clear whether ViT-EE will have the same advantage over CNN-based early exits in ViT backbones as well. Another advantage of using Transformer encoder is the simplicity of its structure, which means it can handle various other data types such as point-clouds and even cross-modal data \cite{2105.15078, 2105.09121}. The main drawback of ViT-EE is its high overhead, however, the building blocks of the recently proposed attention-free MLP-based architectures can serve as more lightweight alternatives that still maintain a global receptive field and structure simplicity, leading to \textit{ResMLP-EE} and \textit{MLP-Mixer-EE}.

\begin{figure}
\setlength{\fboxrule}{0pt}
\begin{center}
\begin{tabular}{ c c }
\fbox{\includegraphics[height=3cm]{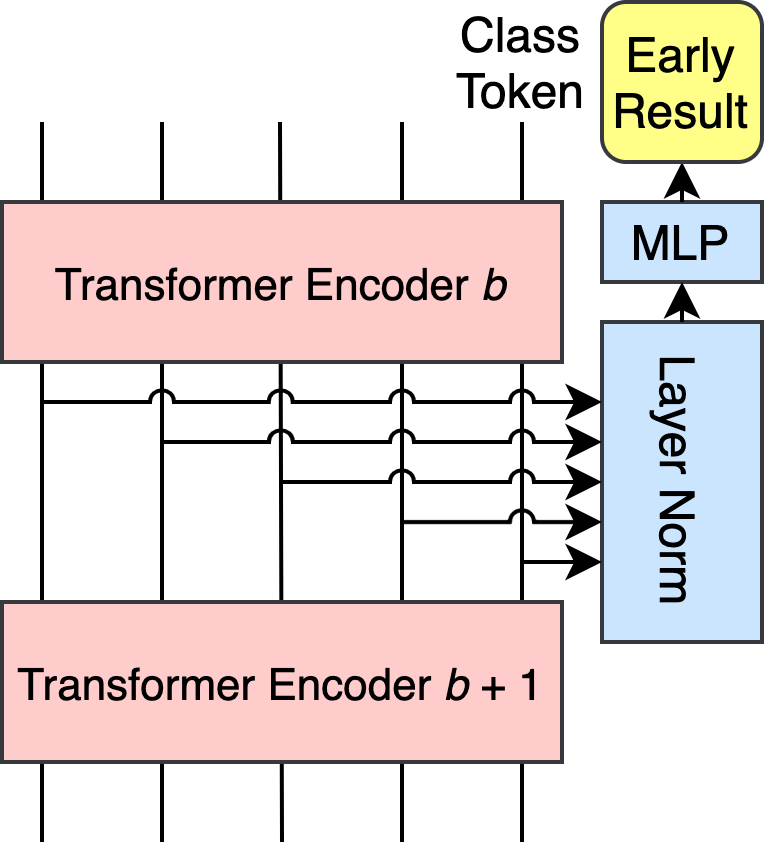}}&
\fbox{\includegraphics[height=3cm]{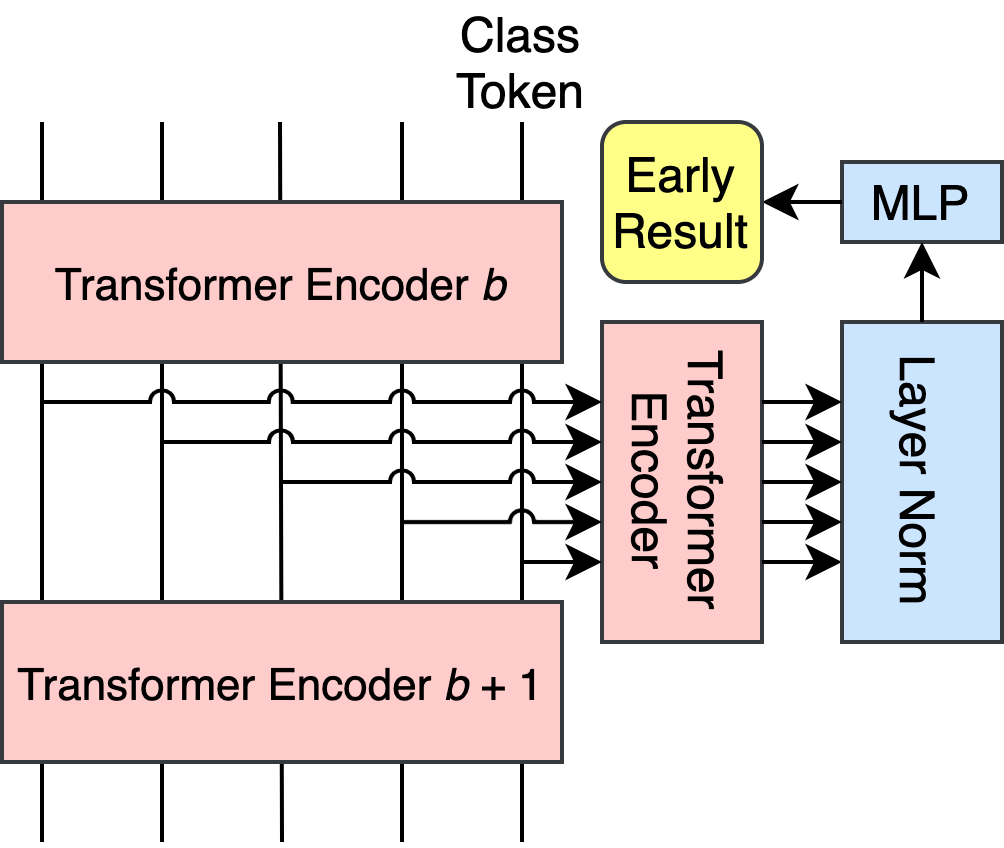}}\\
(a)&(b)\\
\fbox{\includegraphics[height=3cm]{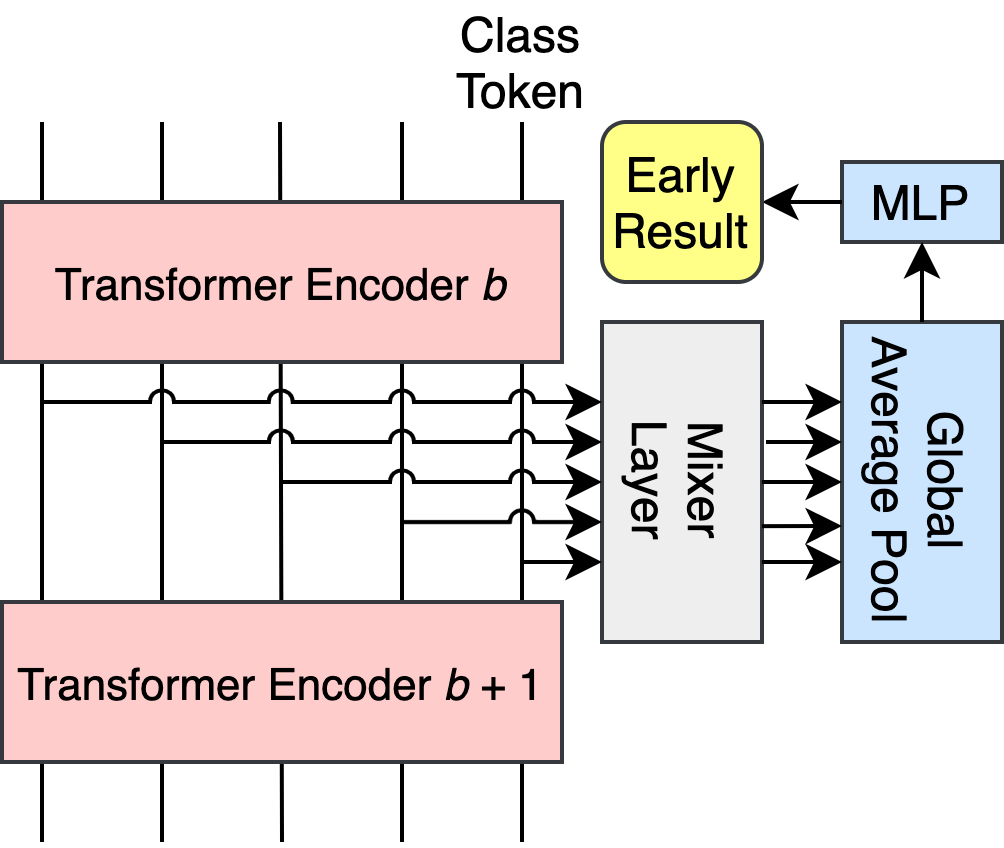}}&
\fbox{\includegraphics[height=3cm]{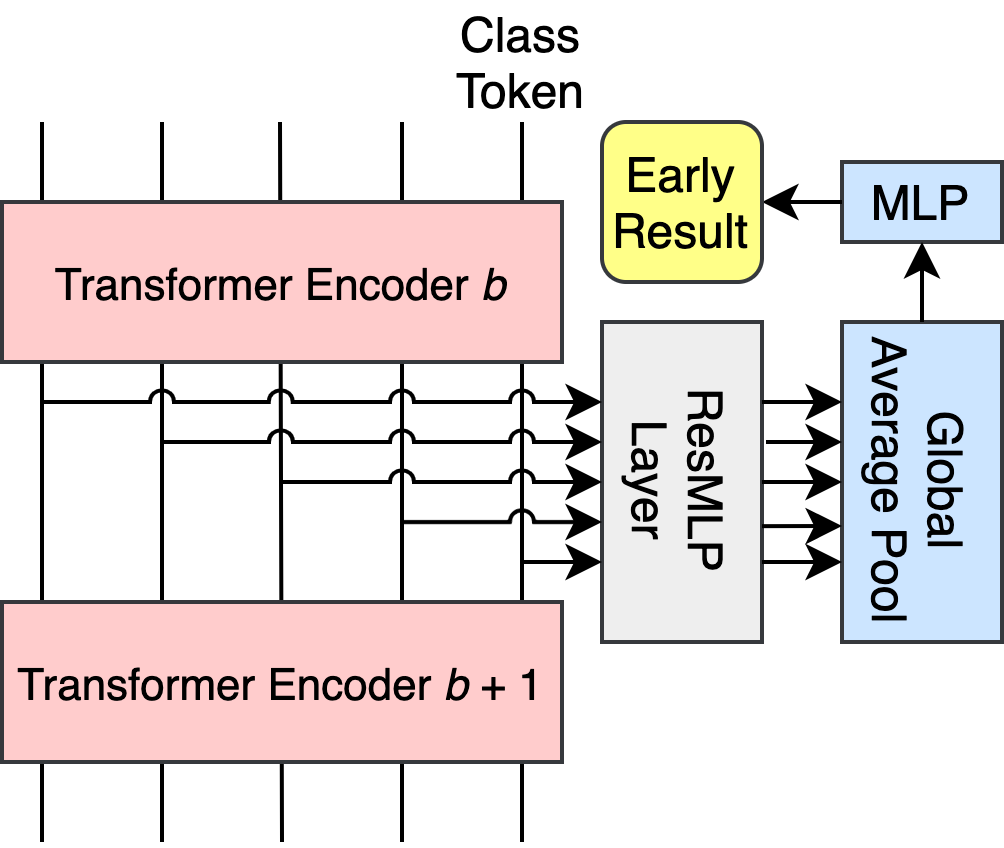}}\\
(c)&(d)
\end{tabular}
\end{center}
\caption{(a) MLP-EE; (b) ViT-EE; (c) MLP-Mixer-EE; and (d) ResMLP-EE early exit branch architectures.}
\label{fig:patch_ee}
\end{figure}

Formally, the output of Transformer encoder $ b $, denoted by $ P^b $, consists of patch embeddings $ p^b_1, \dots, p^b_N $ corresponding to the input image patches, as well vector $ p^b_0 $ corresponding to the classification token. Since the shape of the intermediate representations is the same for all of the hidden layers, without loss of generality, we assume that the early exit branch is to be placed after Transformer encoder $ b $. In MLP-EE, shown in Figure \ref{fig:patch_ee} (a), $ P^b $ is normalized to obtain $ \bar{P}^b = Norm(P^b) $. Subsequently, an MLP consisting of three dense layers with two dropout layers in between takes $ \bar{p}^b_0 $ as input, where $ \bar{P}^b = (\bar{p}^b_0, \dots, \bar{p}^b_N) $, and outputs the early result. The MLP layers in all our proposed architectures have the same three layers. In ViT-EE, shown in Figure \ref{fig:patch_ee} (b), $ P^b $ is given as input to a Transformer encoder layer \cite{2105.09121}. The output of the Transformer encoder is then normalized and passed on to an MLP, similar to the previous architecture.

\begin{figure}
\setlength{\fboxrule}{0pt}
\begin{center}
\begin{tabular}{ c }
\fbox{\includegraphics[height=3cm]{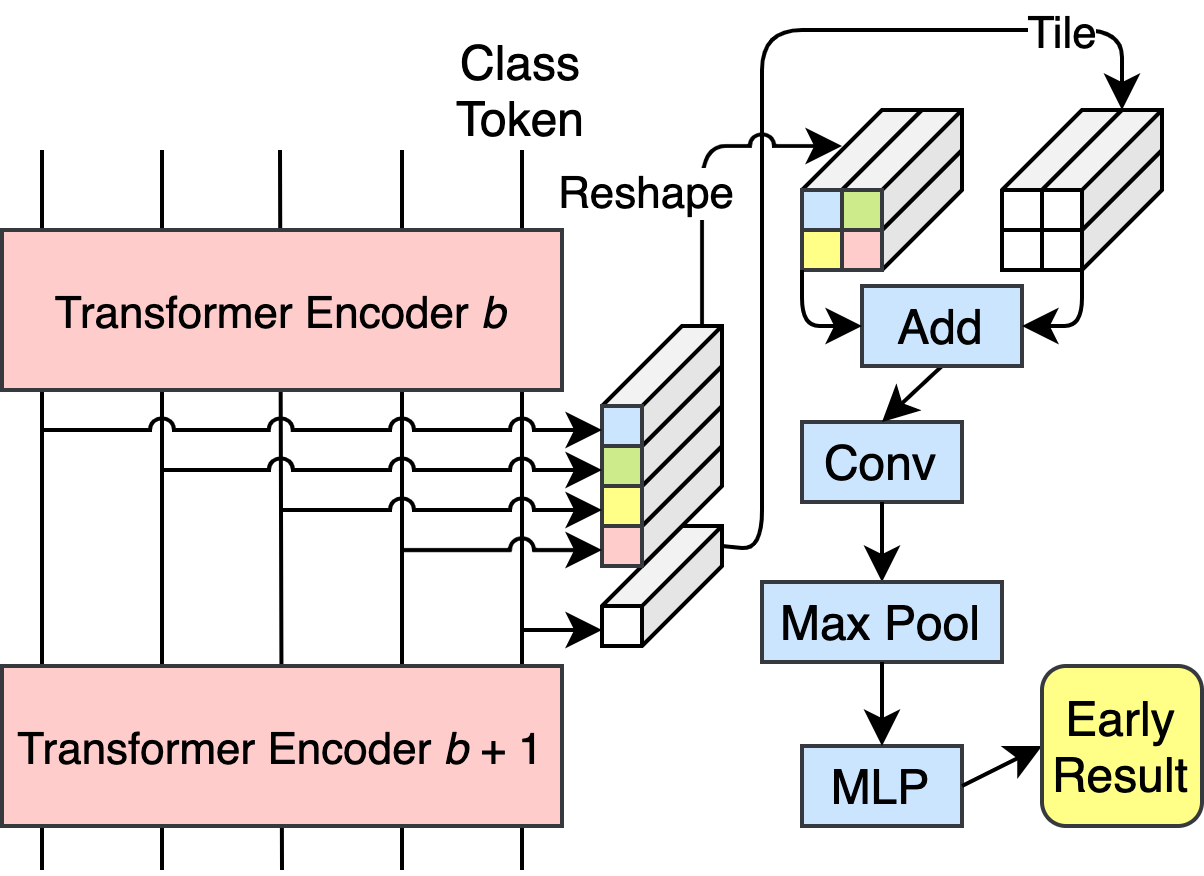}}\\
(a)\\
\fbox{\includegraphics[height=3cm]{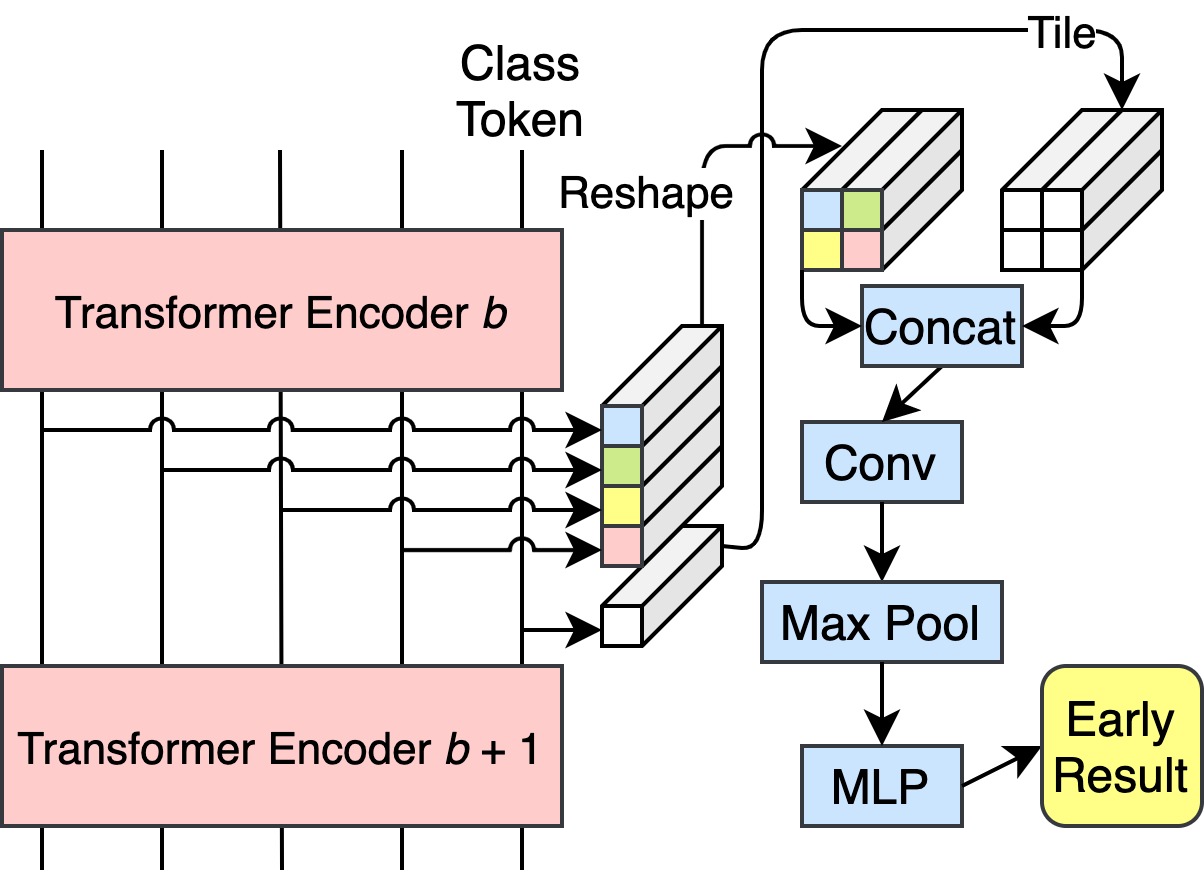}}\\
(b)\\
\fbox{\includegraphics[height=3cm]{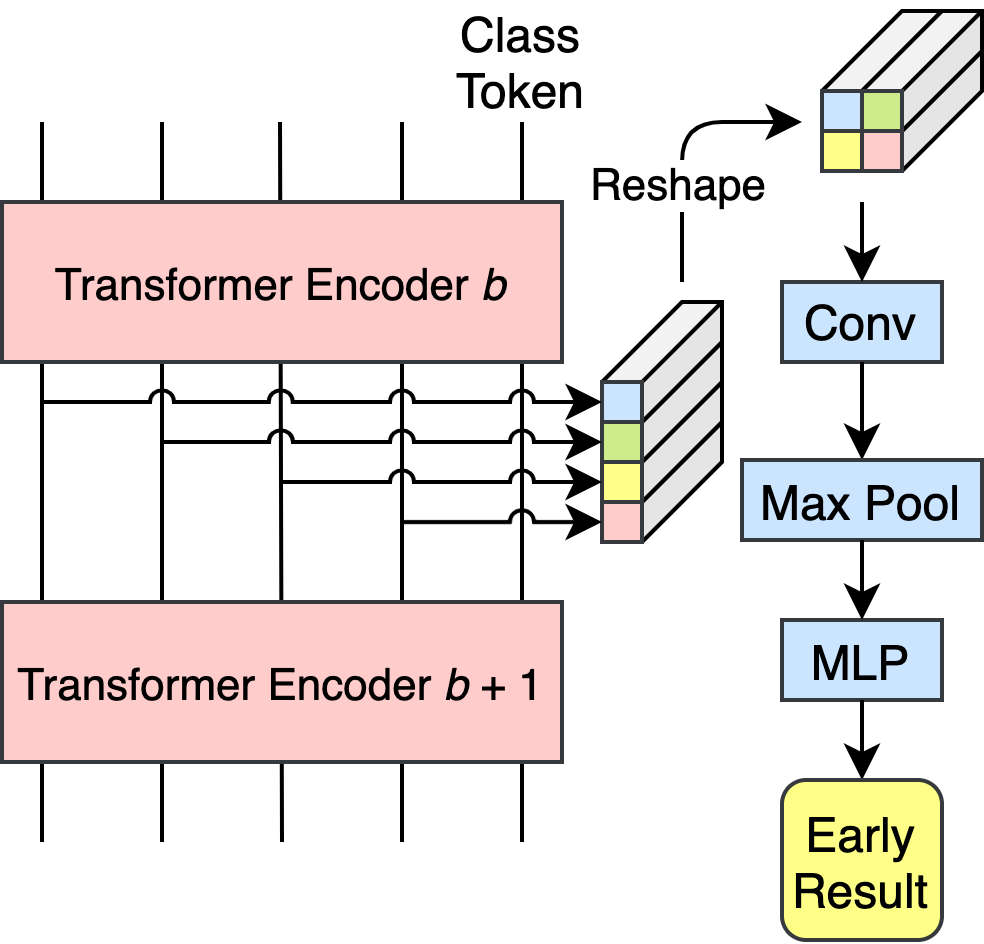}}\\
(c)
\end{tabular}
\end{center}
\caption{(a) CNN-Add-EE; (b) CNN-Project-EE and (c) CNN-Ignore-EE early exit branch architectures.}
\label{fig:cnn_ee}
\end{figure}

In CNN-based early exits, the $ N $ patch embeddings $ p^b_1, \dots, p^b_N $ can be reshaped into a tensor $ C^b \in \mathbb{R}^{\sqrt{N} \times \sqrt{N} \times d_v} $, akin to an intermediate representation in a CNN backbone, with height and width of $ \sqrt{N} $ and $ d_v $ channels, and then passed on to a convolution layer, a max pooling layer and an MLP to obtain the early result. However, it is not clear what should be done with classification token $ \bar{p}^b_0 $. A similar situation arises in dense prediction using Vision Transformers, where three ways for dealing with the classification token are proposed \cite{2103.13413}. In CNN-Add-EE, the classification token is added to every patch embedding, leading to $ \bar{C}^b = (p^b_1 + p^b_0, p^b_2 + p^b_0, \dots, p^b_N + p^b_0) $; in CNN-Project-EE, the classification token is concatenated to every patch embedding, leading to $ \bar{C}^b = (concat(p^b_1, p^b_0), concat(p^b_2, p^b_0), \dots, concat(p^b_N, p^b_0)) $; and in CNN-Ignore-EE, the classification token is ignored and discarded, leading to $ \bar{C}^b = C^b $. These three alternative architectures are depicted in figure \ref{fig:cnn_ee}.

As previously mentioned, the building blocks of attention-free MLP-based architectures can be low-overhead alternatives for ViT-EE which uses a Transformer encoder layer. Figure \ref{fig:patch_ee} (c) and (d) shows the MLP-Mixer-EE and ResMLP-EE early exit branch architectures, respectively. Note that similar to the original MLP-Mixer and ResMLP architectures, the output of the mixer layer and the ResMLP layer are passed on to a global average pooling layer.

\section{Results}
\label{sec:results}

For the image classification experiments, we use CIFAR-10, CIFAR-100 and Fashion MNIST datasets \cite{Krizhevsky2009, 1708.07747}. We use ViT-B/16 architectures with the original pre-trained weights provided by the authors \cite{dosovitskiy2021an} as backbones, and we train them on our target datasets using a cross-entropy loss function. For the regression experiments, we investigate crowd counting, which is the problem of counting the total number of people present in a given image \cite{2003.12783}. We use DISCO \cite{2005.07097} as the dataset and TransCrowd \cite{2104.09116} which is a ViT-based architecture as the backbone. Mean absolute error (MAE) is commonly used to evaluate the accuracy of crowd counting models \cite{1906.09707}. All three backbones have 12 Transformer encoder layers.

All our models were trained using the Adam optimizer \cite{DBLP:journals/corr/KingmaB14} with learning rates of $ \{ 10^{-3} $, $10^{-4}$, $10^{-5} \} $ and the best result is selected. The learning rate is reduced by a factor of 0.6 on plateau with a tolerance of 2 epochs, and an early stopping mechanism with a tolerance of 5 epochs is used.

Note that while early exits have been recently attached to high-performing CNN backbones \cite{2104.10461, 2105.09121, DBLP:conf/icml/HacohenW19}, there is no prior work for early exits on Vision Transformer backbones. Since the performance obtained by Vision Transformer backbones is improved by a large margin, we omit listing the comparison with early exits on CNN backbones in the following results.

\subsection{Classifier-Wise}

\begin{figure*}
\setlength{\fboxrule}{0pt}
\begin{center}
\begin{tabular}{ c c }
(a) & \fbox{\includegraphics[height=5cm]{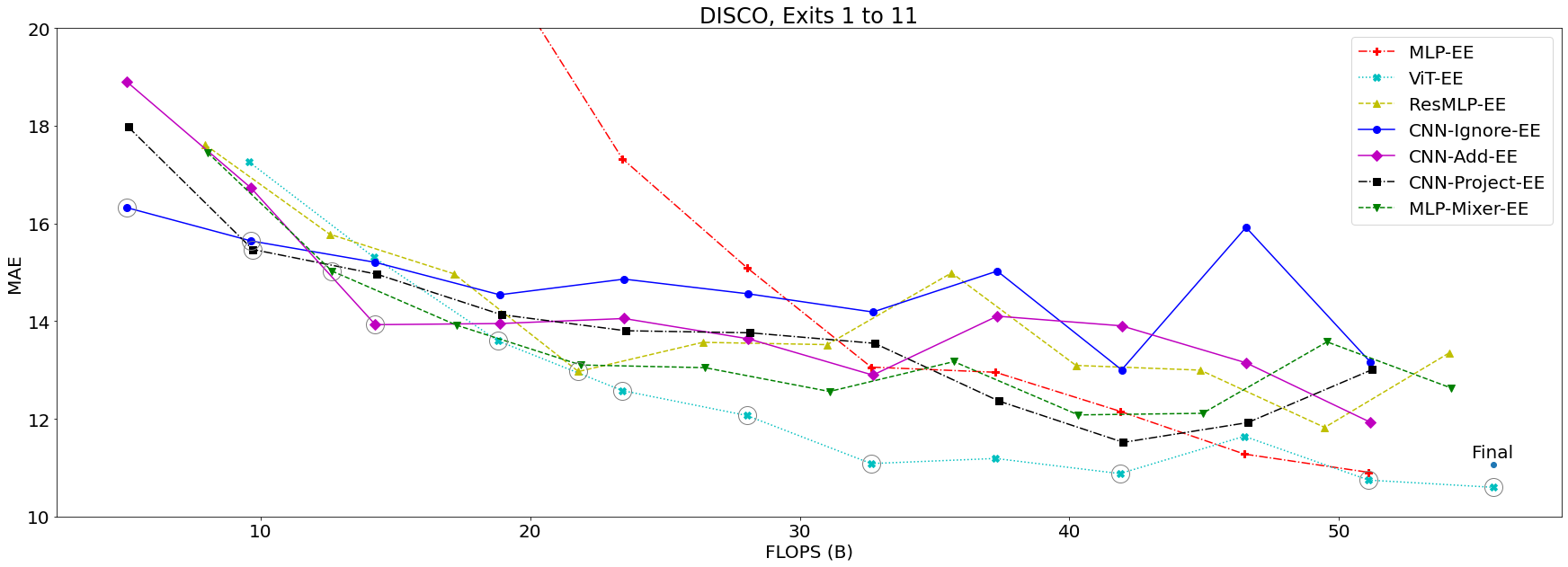}}\\
(b) & \fbox{\includegraphics[height=5cm]{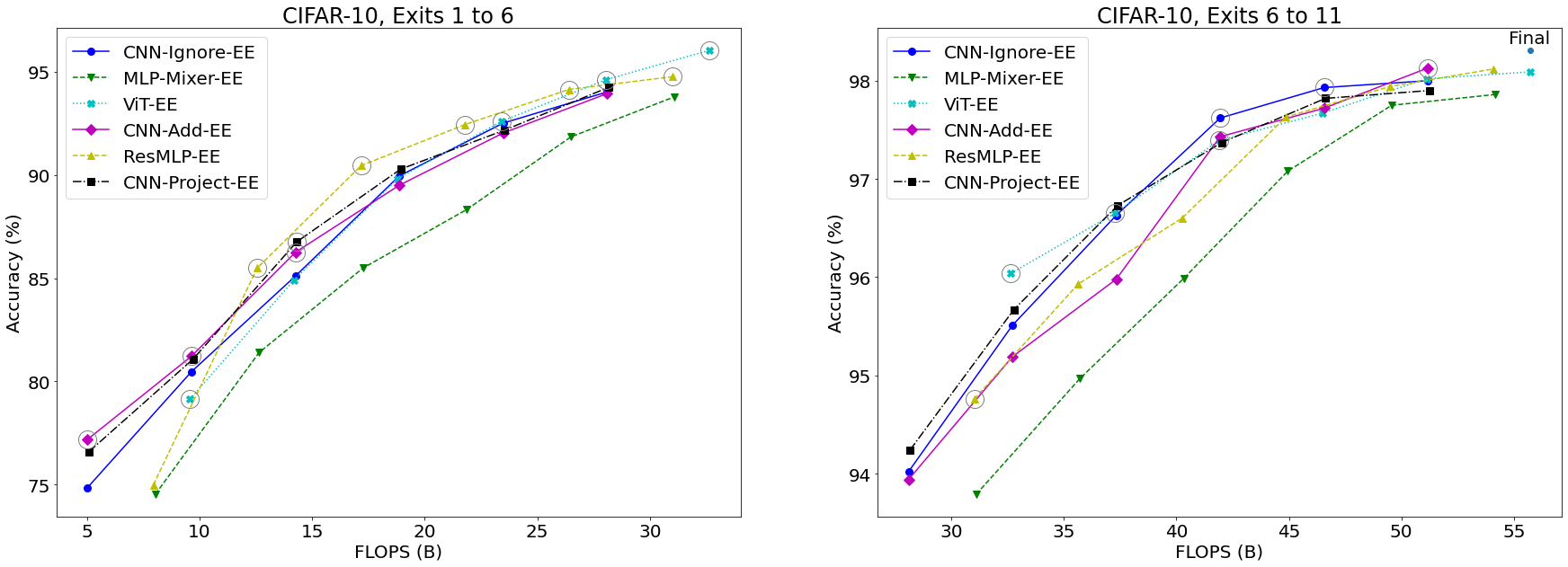}}\\
(c) & \fbox{\includegraphics[height=5cm]{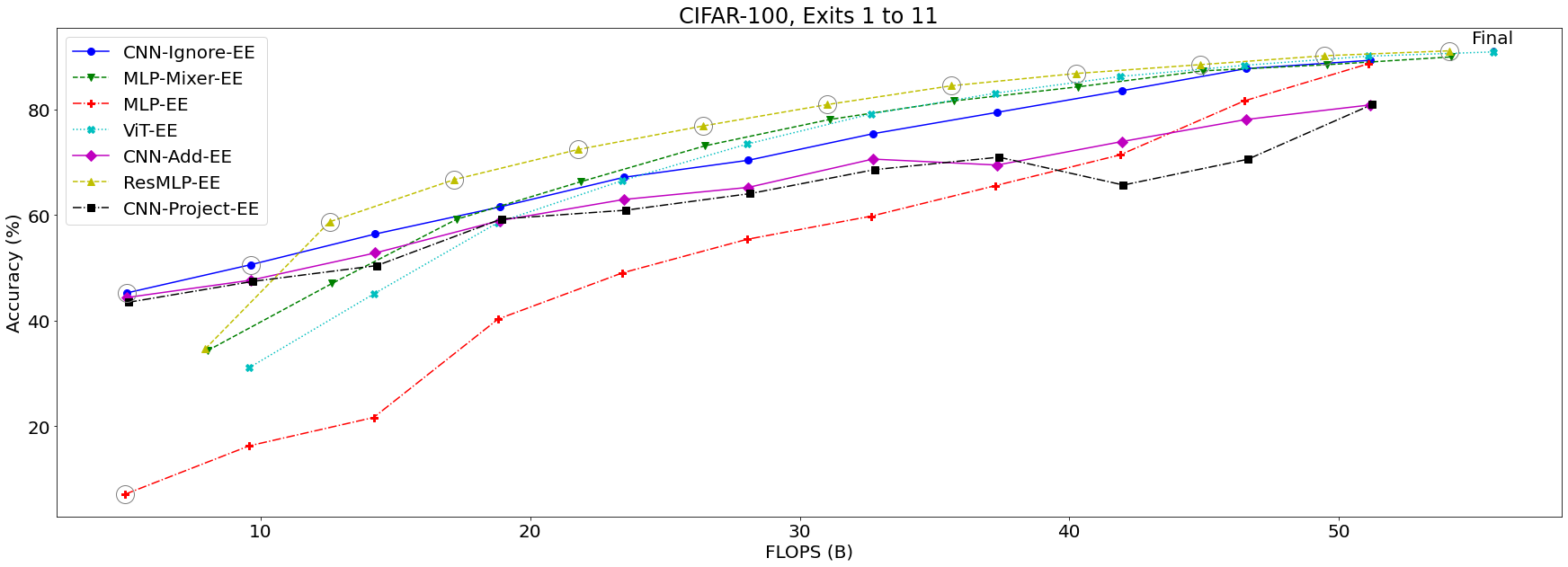}}\\
(d) & \fbox{\includegraphics[height=5cm]{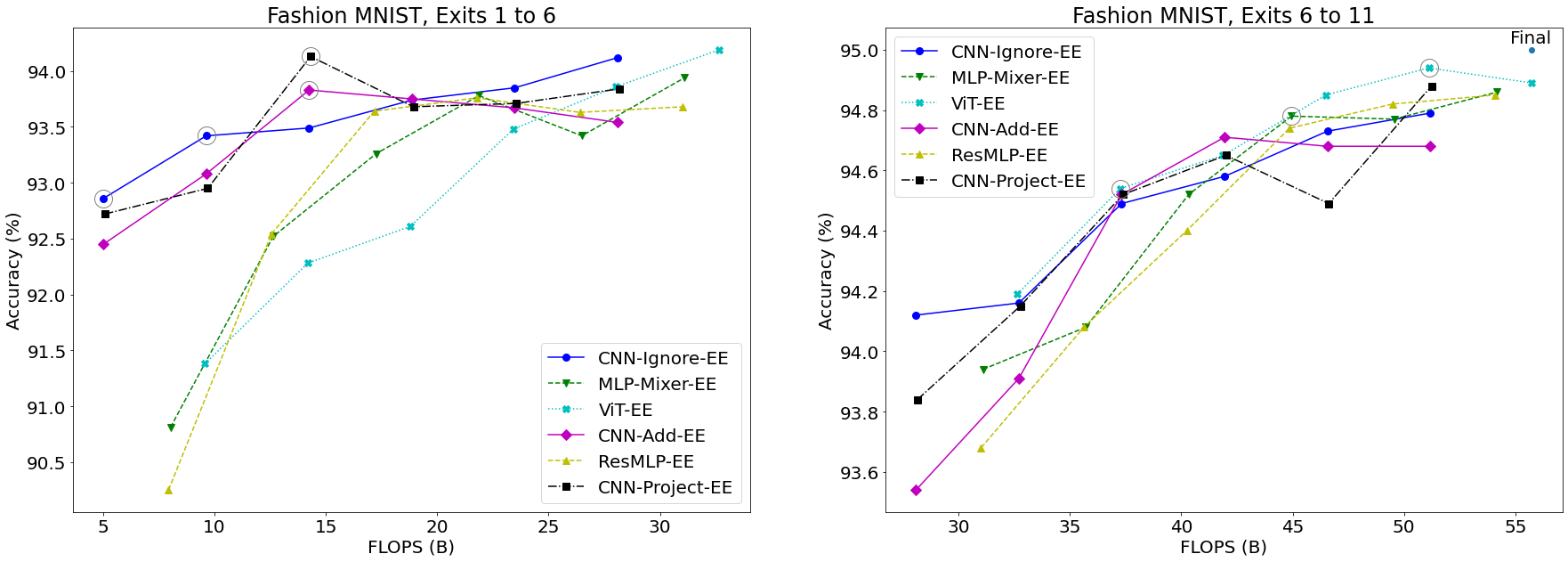}}\\
\end{tabular}
\end{center}
\caption{Performance of different multi-exit architectures on (a) DISCO; (b) CIFAR-10; (c) CIFAR-100 and (d) Fashion MNIST datasets trained with classifier-wise strategy. Practical early exit branches are circled. In order to highlight the differences, early exits with MLP-EE architecture are removed in (b) and (d) and branches 1 to 6 and 6 to 11 are separated.}
\label{fig:cw_all}
\end{figure*}

With the classifier-wise training strategy, since the branches do not affect each other or the backbone, we place and train early exit branches with each of our proposed architectures at every single layer in the backbone. For each early exit branch, we record the accuracy of its early results as well as the total FLOPs up to and including the branch. The results are depicted in Figure \ref{fig:cw_all}, where all practical early exits are circled. In addition, the accuracy of the final exit of the backbone is shown in these figures.

These results can be used to select a collection of lightweight high-performing branches. With dynamic inference, it is desirable for the model to be as fine-grained as possible, therefore, with the classifier-wise strategy where placing more branches does not affect other branches or the backbone, it is desirable to place as many branches as possible on the backbone. To make this more clear, imagine a scenario where only two exit options are available: option A with 80\% accuracy and 3B FLOPS, and option B with 90\% accuracy and 6B FLOPS. If the computation budget at hand is 5B FLOPS, the only possible option to choose is A, resulting in 80\% accuracy. However, with a finer-grained model that also includes option C with 85\% accuracy and 4B FLOPS, choosing option C leads to 85\% accuracy. Hence we examine all possible branch locations: if there exists a single practical branch at a location, that branch should be added at that location; if there are no practical branches at a location, then no branches should be added there, since  more accurate and more lightweight exits are available; and if there are more than one practical branches at a location (for instance, with the DISCO dataset in Figure \ref{fig:cw_all} (a), both CNN-Ignore-EE and CNN-Project-EE make practical branches at layer 2) it means that there is a trade-off between accuracy and computation at that location, and the proper branch should be selected based on the particular application. Alternatively, it is possible to deploy multiple branches at the same location simultaneously, and exit the one that fits the budget during inference. Note that with the classifier-wise strategy, there can be different branch types on the same backbone, for instance, there can be a CNN-Add-EE branch at location 1 and a ViT-EE branch at location 2.

Several observations can be made from these results. First, all of our proposed architectures create at least one practical branch. As expected, MLP-EE does not contain enough parameters and layers to extract useful features in early locations on its own, and thus performs poorly, while it catches up in the later locations where the features extracted by the intermediate layers can compensate. Furthermore, MLP-EE only processes the classification head, which contains only low-level features in very early layers. However, MLP-EE always creates the first practical branch as it is the most lightweight. Secondly, CNN-based branches outperform other types in the first few locations. This is likely because the fusion of convolutional layers that capture local interactions well, with the global attention of the backbone, combines the best of both worlds. However, this effect seems to fade in later locations, perhaps since several layers of the backbone are able to capture both local and global interactions fairly well. In addition, CNN-Ignore-EE outperforms other CNN-based early exits in most of these early cases, as the classification token in the very early layers contains only low-level features. Thirdly, aside from the very early locations where CNN-based branches dominate, ResMLP-EE performs better in classification problems, while ViT-EE performs better in crowd counting. Evident from the use of visual attention mechanisms and dilated convolutions in many high-performing models for crowd counting \cite{2003.12783}, global information such as perspective plays an important role in crowd counting, therefore, ViT-EE which can capture multiple types of attention through the use of the multi-head attention layer in Transformer encoder, outperforms ResMLP-EE which does not include a mechanism for handling multiple types of attention. Fourthly, observe that MLP-Mixer-EE outperforms ResMLP-EE in most locations in the crowd counting cases. This is because the affine transformation in ResMLP can be used instead of normalization when the training is stable \cite{2105.03404}, however, with crowd counting, the training process is not as stable as image classification.

Note that in the last six locations in CIFAR-10 and Fashion MNIST cases, the differences between the performance of different branch types are minuscule, and therefore less informative. Moreover, observe that unlike multi-exit architectures with CNN backbones, branches placed later on a ViT backbone do not necessarily provide a higher accuracy compared to previous branches. This is because in CNN backbones, the network has a very local receptive field in the early layers, and the receptive field gradually increases throughout the network, whereas ViTs have a global receptive field from the very first layer. This means that the accuracy of later branches of CNN backbones is expected to increase since the receptive field has increased, whereas in ViT backbones, later intermediate layers do not have any advantages in terms of the receptive field, thus their branches may obtain a lower accuracy. Finally, note that in the DISCO experiments, some of the very late early exit branches achieve a lower MAE compared to the final exit. This is because the MLP in our proposed architectures consists of three layers while the MLP in the ViT-B/16 backbone has one.

\subsection{End-to-End and Layer-Wise}

Unlike the classifier-wise training strategy, it is not possible to conduct a comprehensive study of the end-to-end and layer-wise strategies, since there are $ 2^{L - 1} - 1 $ possible branch placements. In addition, the end-to-end strategy can have infinitely many weight values for each of the placements. Therefore, for these training strategies, we only investigate two cases; one where a single early exit branch is placed after the sixth layer; the other where three branches are placed after the third, sixth and ninth layers. In both cases, the contribution of the final exit to the loss is double the contribution of the early exits.

Results are summarized in Tables \ref{tab:ee_1_exit} and \ref{tab:ee_3_exits}. In all image classification cases, final accuracy is decreased compared to the backbones without early exits, which have an accuracy of 98.31\% for CIFAR-10, 91.24\% for CIFAR-100 and 95.00\% for Fashion MNIST. However, in crowd counting, the final MAE is improved from the original 11.07 when a single MLP-EE, ViT-EE or MLP-Mixer-EE branch is used. Similar to the classifier-wise strategy, in both cases involving the DISCO dataset, MLP-Mixer-EE outperforms ResMLP-EE for the same reason explained above. Furthermore, ViT-EE outperforms other branch types in most cases, particularly when there are 3 exit branches, and performs very high in others. Since the end-to-end training strategy affects the parameters of the backbone, perhaps ViT-EE branches have the least negative impact on the backbone due to the similarity of their architecture with the layers of the backbone. This is further supported by the fact that CNN-based branches whose architectures differ the most from that of the backbone, typically perform much worse.

\begin{table}
\begin{center}
\resizebox{8.9cm}{!}{
\begin{tabular}{| c | c c | c c | c c | c c | c |}
\hline

Branch Arch. &
\multicolumn{2}{c}{CIFAR-10 Acc.} &
\multicolumn{2}{c}{CIFAR-100 Acc.} & 
\multicolumn{2}{c}{Fasion MNIST Acc.} & 
\multicolumn{2}{c}{DISCO MAE} &
FLOPS\\

&
Early@6& 
Final&
Early@6& 
Final&
Early@6& 
Final&
\\

\hline\hline

MLP-EE&
94.90\%&
96.73\%&
81.12\%&
87.45\%&
\textbf{94.47\%}&
94.85\%&
\textbf{11.04}&
\textbf{10.72}&
28.04\\

CNN-Ignore-EE&
\textbf{95.95\%}&
97.10\%&
77.97\%&
85.90\%&
94.43\%&
94.67\%&
21.88&
11.09&
28.10\\

CNN-Add-EE&
94.94\%&
96.87\%&
75.75\%&
86.96\%&
94.22\%&
94.77\%&
18.46&
11.24&
28.10\\

CNN-Project-EE&
94.66\%&
96.80\%&
77.95\%&
86.89\%&
94.33\%&
94.69\%&
18.23&
11.29&
28.16\\

ViT-EE&
95.89\%&
96.99\%&
\textbf{85.23\%}&
\textbf{89.44\%}&
94.39\%&
94.84\%&
11.06&
11.01&
32.65\\

MLP-Mixer-EE&
95.78\%&
97.07\%&
81.72\%&
87.53\%&
94.41\%&
94.88\%&
13.03&
10.93&
31.11\\

ResMLP-EE&
95.44\%&
\textbf{97.35\%}&
82.41\%&
87.57\%&
94.38\%&
\textbf{94.95\%}&
16.99&
11.36&
31.02\\

\hline
\end{tabular}
}
\end{center}
\caption{Performance of multi-exit architectures with one branch, trained with end-to-end strategy. The last column shows the FLOPS up to and including the branch.}
\label{tab:ee_1_exit}
\end{table}

\begin{table}
\begin{center}
\resizebox{\columnwidth}{!}{
\begin{tabular}{| c | c c c c | c c c c | c c c c |}
\hline

Branch Arch. &
\multicolumn{4}{c}{CIFAR-10 Acc.} &
\multicolumn{4}{c}{CIFAR-100 Acc.} & 
\multicolumn{4}{c}{DISCO MAE}\\

&
Early@3&
Early@6&
Early@9&
Final&
Early@3&
Early@6&
Early@9&
Final&
Early@3&
Early@6&
Early@9&
Final\\

\hline\hline

MLP-EE&
87.21\%&
94.48\%&
95.64\%&
96.19\%&
61.00\%&
79.83\%&
84.42\%&
86.46\%&
13.77&
11.54&
\textbf{11.55}&
11.44\\

CNN-Ignore-EE&
91.44\%&
95.68\%&
96.55\%&
96.56\%&
65.08\%&
79.35\%&
84.74\%&
86.32\%&
20.99&
23.65&
20.88&
11.42\\

CNN-Add-EE&
90.27\%&
95.63\%&
96.80\%&
96.94\%&
62.66\%&
78.86\%&
85.11\%&
87.01\%&
18.33&
19.25&
18.92&
11.77\\

CNN-Project-EE&
91.19\%&
95.77\%&
96.81\%&
96.99\%&
64.26\%&
78.63\%&
84.47\%&
86.19\%&
21.25&
21.46&
17.99&
11.49\\

ViT-EE&
92.35\%&
\textbf{96.01\%}&
\textbf{97.25\%}&
\textbf{97.33\%}&
\textbf{74.73\%}&
\textbf{84.31\%}&
\textbf{87.43\%}&
\textbf{87.88\%}&
12.76&
\textbf{11.27}&
11.59&
11.18\\

MLP-Mixer-EE&
91.16\%&
95.99\%&
96.96\%&
96.99\%&
66.24\%&
81.84\%&
86.68\%&
87.29\%&
\textbf{12.24}&
13.95&
15.29&
11.46\\

ResMLP-EE&
\textbf{92.53\%}&
95.87\%&
96.76\%&
96.86\%&
70.45\%&
82.61\%&
87.36\%&
87.85\%&
14.15&
14.71&
17.05&
\textbf{11.09}\\

\hline
\end{tabular}
}
\end{center}
\caption{Performance of multi-exit architectures with 3 branches, trained with end-to-end strategy.}
\label{tab:ee_3_exits}
\end{table}

With the layer-wise strategy, we encounter a problem. Since Vision Transformers are data-hungry \cite{2101.01169}, they need to be pre-trained on very large datasets. For the first step of the layer-wise strategy where all layers up to and including the first early exit branch are trained, pre-trained weights exist, therefore, the training procedure achieves results with a high accuracy on par with their end-to-end counterpart. For instance, in the case of CIFAR-10, CNN-Ignore-EE achieves 95.97\% accuracy at the sixth layer. However, for subsequent steps, the original pre-trained weights can not be used since the weights of earlier layers have changed. We tested the training process with no pre-trained weights, with the original pre-trained weights as well as pre-trained weights from the end-to-end strategy, however, neither achieved a high accuracy.

\section{Conclusion and Future Direction}
\label{sec:conclusion}

We proposed seven architectures for early exiting Vision Transformer backbones, provided the motivations behind each of these architectures, and showed that depending on the branch locations, training strategy and the problem at hand, any of our proposed architectures can prove useful. Furthermore, we provided recommendations on selecting lightweight high-performing branches based on the results of our experiments. We discussed the role of architectural elements such as local and global interactions, receptive field, classification token, support for multiple types of attention, normalization and similarity of branch architecture with the backbone layers, on the performance of multi-exit ViT architectures.

A potential future research direction would be to check whether other recent architectures for computer vision that operate on a sequence of image patches such as Perceiver \cite{2103.03206}, gMLP \cite{2105.08050} and FNet \cite{2105.03824} produce suitable early exits for Vision Transformer backbones.

\section*{Acknowledgment}

This work was funded by the European Union’s Horizon 2020 research and innovation programme under grant agreement No 957337, and by the Danish Council for Independent Research under Grant No. 9131-00119B. This publication reflects the authors views only. The European Commission and the Danish Council for Independent Research are not responsible for any use that may be made of the information it contains.

\bibliographystyle{IEEEtran}
\bibliography{references.bib}

\end{document}